\documentclass[final]{cvpr}

\usepackage{times}
\usepackage{epsfig}
\usepackage{graphicx}
\usepackage{amsmath}
\usepackage{amssymb}

\usepackage{xcolor}
\usepackage{multirow}
\usepackage{booktabs}
\usepackage{algorithm, algorithmic}
\usepackage{fancyhdr}

\usepackage[pagebackref=true,breaklinks=true,colorlinks,bookmarks=false]{hyperref}

\def\cvprPaperID{22} 
\def\confYear{CL Workshop 2021}

\newcommand{\imagenetoverhead}[1]{
\begin{figure}[t]
    \centering
    \includegraphics[trim=20 0 20 0, clip, width=\linewidth]{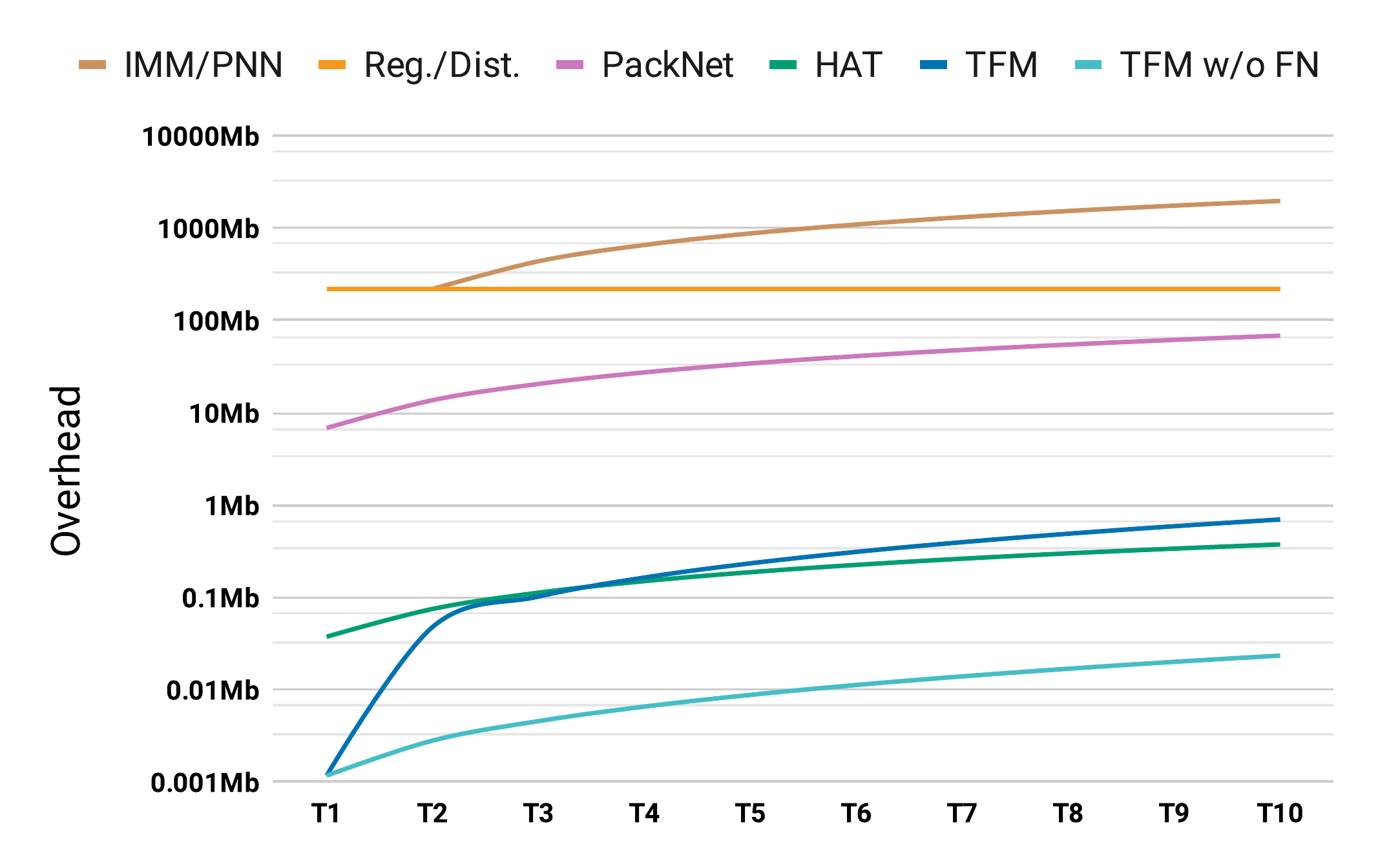}
    \caption{#1}
    \label{fig:overhead}
\end{figure}
}
\newcommand{\soacomparisontable}{
\begin{table*}
\setlength\tabcolsep{5pt}
\caption{Summary of related work characteristics. \textit{pf}: per feature, \textit{pp}: per parameter, \textit{pt}: per task, \textit{pfp}: per feature and parameter.}
\label{tab:soa_comparison}
\vspace{0.2cm}
\centering
\resizebox{\linewidth}{!}{
\begin{tabular}{ccccccccc}
\toprule
\multirow{ 2}{*}{\textbf{Family}} & \multirow{ 2}{*}{\textbf{Method}} & \multirow{ 2}{*}{\shortstack{\textbf{Revisit}\\\textbf{data}}} & \multirow{ 2}{*}{\shortstack{\textbf{Require}\\\textbf{Backbone net}}} & \multirow{ 2}{*}{\shortstack{\textbf{Easily}\\\textbf{expandable}}} & \multirow{ 2}{*}{\textbf{Overhead}} & \multirow{ 2}{*}{\textbf{Forgetting}} & \multirow{ 2}{*}{\shortstack{\textbf{Forward}\\\textbf{transfer}}} & \multirow{ 2}{*}{\shortstack{\textbf{Features}\\\textbf{or weights}}}\\
& & & & & & & & \\
\midrule
\multirow{ 3}{*}{Baseline} & Finetune & No & No & Yes & None & Yes & Little & neither \\
 & Joint & Yes & No & Yes & None & No & Little & neither \\
 & Freeze & No & Yes & Yes & None & No & Backbone only & neither \\
\midrule
\multirow{ 4}{*}{Distillation} & LwF~\cite{li2017learning}      & No  & No & No  & 1 float \textit{pp}   & Some & Yes & weights \\
                              & LFL~\cite{jung2016less}        & No  & No & No  & 1 float \textit{pp}   & Some & Yes & weights \\
                              & PNN~\cite{rusu2016progressive} & No  & No & Yes & duplicate \textit{pt} & Some & Yes & weights \\
                              & P\&C~\cite{schwarz2018progress} & No  & No & No & extra network & Some & Little & weights \\
\midrule
\multirow{ 6}{*}{Model-based} & EWC~\cite{kirkpatrick2017overcoming} & No & No & No & 1 float \textit{pp}    & Some & Yes & weights \\
                              & R-EWC~\cite{liu2018rotate}           & No & No & No & 1 float \textit{pp}    & Some & Yes & weights \\
                              & IMM~\cite{lee2017overcoming}         & No & No & No & 1 float \textit{pp pt} & Some & Yes & weights \\
                              & SI~\cite{zenke2017continual}         & No & No & No & 1 float \textit{pp}    & Some & Yes & weights \\
                              & MAS~\cite{aljundi2018memory}         & No & No & No & 1 float \textit{pp}    & Some & Yes & weights \\
                              & SSL~\cite{aljundi2018selfless}       & No & No & No & 1 float \textit{pfp}   & Some & Yes & both \\
\midrule
\multirow{ 5}{*}{Mask-based} & PackNet~\cite{mallya2018packnet}     & No & No & No  & 1 int \textit{pp pt}    & No & Yes & weights \\
                             & PiggyBack~\cite{mallya2018piggyback} & No & Yes & No  & 1 bit \textit{pp pt}   & No & Backbone only & weights \\
                             & HAT~\cite{serra2018overcoming}       & No & No  & Yes  & 1 float \textit{pf pt} & Some & Yes & features \\
                             & TFM w/o FN (Ours)                    & No & No  & Yes & 2 bits \textit{pf pt}  & No & Yes & features \\
                             & TFM (Ours)                           & No & No  & Yes & 2 bits + 2 floats \textit{pf pt} & No & Yes & features \\
\bottomrule
\end{tabular}}
\end{table*}}
\newcommand{\weightsfeatures}[1]{
\begin{table}
\caption{#1}
\label{table:networks}
\vspace{0.1cm}
\centering
\resizebox{0.75\linewidth}{!}{
\begin{tabular}{ccc}
\toprule
\textbf{Network} & \textbf{\#weights} & \textbf{\#features} \\
\midrule
LeNet~\cite{lecun1998gradient} & 59,956 & 226 \\
AlexNet~\cite{krizhevsky2012imagenet} & 54,547,712 & 9,344 \\
VGGNet~\cite{simonyan2014very} & 119,579,904 & 10,880 \\
ResNet-50~\cite{he2016deep} & 19,330,304 & 22,720 \\
\bottomrule
\end{tabular}}
\vspace{-0.2cm}
\end{table}
}
\newcommand{\figbinarymask}{
\begin{figure}[t!]
    \centering
    \includegraphics[width=0.95\columnwidth]{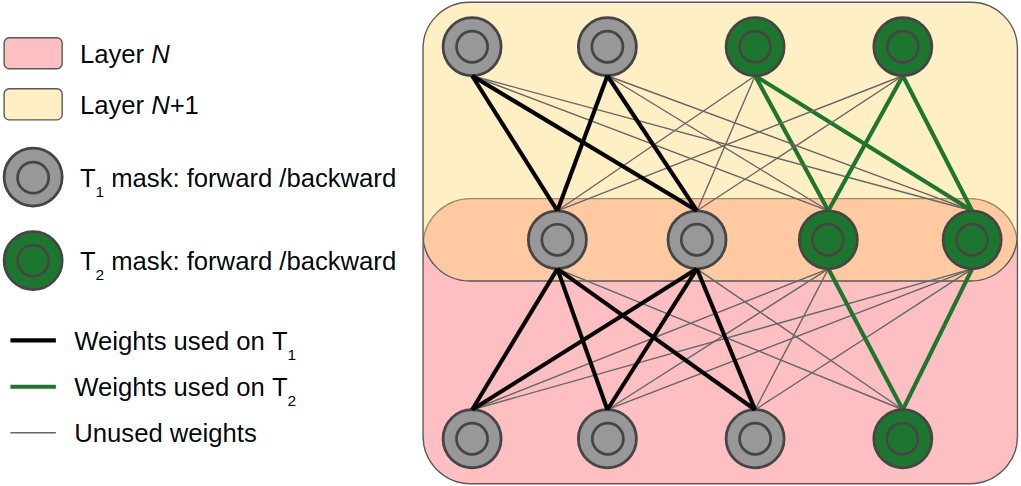}
    \caption{Binary masks encode two states: used or unused. In this case, neurons in grey are learnable for task 1 but neurons in green are not, and the opposite is true for task 2. All grey weights are unused by both tasks.}
    \label{fig:binary_mask}
\end{figure}
}
\newcommand{\figternarymask}{
\begin{figure}[t!]
    \centering
    \includegraphics[width=0.95\columnwidth]{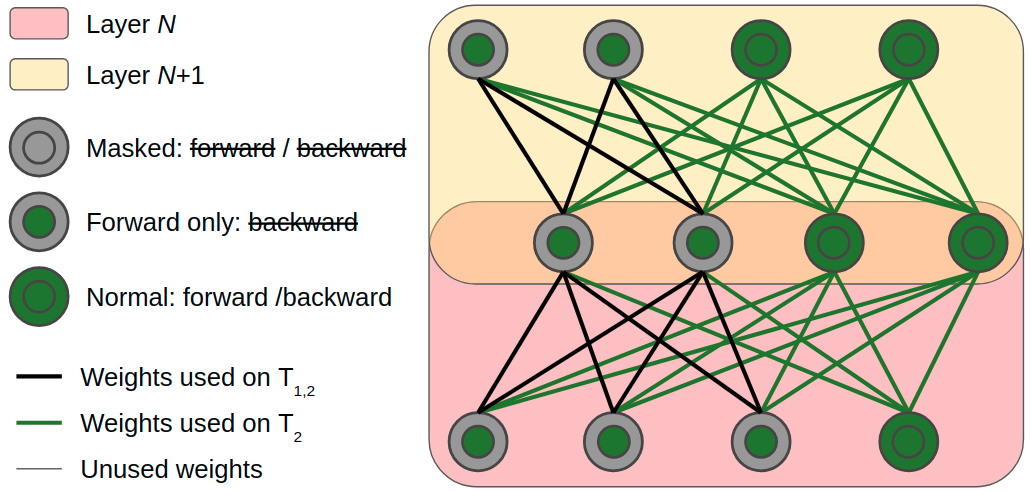}
    \caption{Ternary masks encode three states: masked (frozen), forward only or normal (forward and backward). In comparison to Fig.~\ref{fig:binary_mask}, all unused connections can now be learned without forgetting previous knowledge.}
    \label{fig:ternary_mask}
\end{figure}
}
\newcommand{\growingalgorithm}{
\begin{algorithm}[t]
\caption{: Growing Ternary Feature Masks}
\label{table:algorithm}
\begin{algorithmic}
\STATE\textbf{Input:} ternary mask $m$ for each task and layer \\
\STATE\textbf{Input:} $s$ number of features to add per layer \\
\STATE\textbf{Require:} Network with $L$ layers \\
\STATE\textbf{Require:} Tasks $1,\ldots,T$, current task $t > 1$ \\
\commentgrey{\% Loop for each layer \%} \\
\textbf{for} $l=1,\ldots,L$ \\
\ \ \ \ old\_size $\leftarrow$ current\_output\_size($l$) \\
\ \ \ \ new\_size $\leftarrow$ old\_size + $s^{l}$ \\
\ \ \ \ \commentgrey{\% For each previous task \%} \\
\ \ \ \ \textbf{for} $k=1,\ldots,t-1$ \\
\ \ \ \ \ \ \ \ $m^{k,l}$[old\_size:new\_size] $\leftarrow$ 0 \\
\ \ \ \ \textbf{end for} \\
\ \ \ \ \commentgrey{\% For the current task \%} \\
\ \ \ \ $m^{t,l}$[0:old\_size] $\leftarrow$ 1 \\
\ \ \ \ $m^{t,l}$[old\_size:new\_size] $\leftarrow$ 2 \\
\textbf{end for} \\
\end{algorithmic}
\end{algorithm}
}
\newcommand{\tabledatasets}{
\begin{table}
\setlength\tabcolsep{5pt}
\caption{Summary of datasets used.}
\label{table:datasets}
\vspace{0.1cm}
\centering
\resizebox{0.9\linewidth}{!}{
\begin{tabular}{ccccc}
\toprule
\textbf{Dataset} & \textbf{\#Train} & \textbf{\#Eval} & \textbf{\#Classes} \\
\midrule
tiny ImageNet~\cite{deng2009imagenet} & 100,000 & 10,000 & 200 \\
ImageNet~\cite{russakovsky2015imagenet} & 1,281,167 & 50,000 & 1000 \\
Oxford Flowers~\cite{nilsback2008automated} & 2,040 & 6,149 & 102 \\
CUB Birds~\cite{wah2011caltech} & 5,994 & 5,794 & 200 \\
Stanford Actions~\cite{yao2011human} & 4,000 & 5,532 & 40 \\
\bottomrule
\end{tabular}}
\end{table}
}
\newcommand{\finegrainedtable}[1]{
\begin{table}
\setlength\tabcolsep{4pt}
\caption{#1}
\label{tab:fine_grained}
\vspace{0.1cm}
\centering
\resizebox{\linewidth}{!}{
\begin{tabular}{cccccc}
\toprule
\multicolumn{6}{c}{\textbf{Oxford 102 Flowers}} \\
Method & Task 1 & Task 2 & Task 3 & Task 4 & Avg. \\
\midrule
Finetuning                           & 10.0 (-20.3) & 5.1 (-17.1) & 6.7 (-13.6) & 17.3 (0.0) & 9.8 \\
Freezing                             & 30.3 (0.0) & 39.8 (0.0) & 32.0 (0.0) & 33.1 (0.0) & 33.8 \\
Joint                                & 54.6 (+24.3) & 58.9 (+11.5) & 57.7 (+4.5) & 47.0 (0.0) & 54.6 \\
\midrule
EWC~\cite{kirkpatrick2017overcoming} & 12.1 (-18.2) & 11.6 (-38.1) & 9.3 (-24.4) & 25.8 (0.0) & 14.7 \\
HAT~\cite{serra2018overcoming}       & 17.2 (-12.7) & 19.3 (-28.5) & 28.6 (+1.4) & 31.6 (0.0) & 24.2 \\
PackNet~\cite{mallya2018packnet}     & 32.0 (0.0) & 53.7 (0.0) & 43.6 (0.0) & 37.9 (0.0) & 41.8 \\
TFM w/o FN                           & 36.4 (0.0) & 54.1 (0.0) & 38.6 (0.0) & 39.0 (0.0) & 42.0 \\
TFM                                  & 36.4 (0.0) & 53.8 (0.0) & 45.5 (0.0) & 37.6 (0.0) & \textbf{43.3} \\
\bottomrule
\toprule
\multicolumn{6}{c}{\textbf{CUBS 200 Birds}} \\
Method & Task 1 & Task 2 & Task 3 & Task 4 & Avg. \\
\midrule
Finetuning                           & 7.4 (-30.2) &  2.6 (-30.0) & 29.7 (-3.4)  & 43.1 (0.0) & 20.7 \\
Freezing                             & 37.6 (0.0) & 35.1 (0.0) & 35.4 (0.0) & 38.4 (0.0) & 36.6 \\
Joint                                & 48.7 (+11.1) & 52.1 (+6.0) & 50.7 (+1.5) & 51.9 (0.0) & 50.8 \\
\midrule
EWC~\cite{kirkpatrick2017overcoming} & 16.2 (-21.4) & 19.0 (-21.2) & 24.2 (-14.0) & 41.7 (0.0) & 25.3 \\
HAT~\cite{serra2018overcoming}       & 18.7 (-1.8)  & 19.4 (-0.4)  & 28.5 (-0.6)  & 31.2 (0.0) & 24.4 \\
PackNet~\cite{mallya2018packnet}     & 35.3 (0.0)   & 42.8 (0.0)   & 44.4 (0.0)   & 45.9 (0.0) & 42.1 \\
TFM w/o FN                              & 42.9 (0.0)   & 44.1 (0.0)   & 48.3 (0.0)   & 49.1 (0.0) & 46.1 \\
TFM                                  & 42.9 (0.0)   & 43.1 (0.0)   & 49.9 (0.0)   & 48.8 (0.0) & \textbf{46.2} \\
\bottomrule
\toprule
\multicolumn{6}{c}{\textbf{Stanford 40 Actions}} \\
Method & Task 1 & Task 2 & Task 3 & Task 4 & Avg. \\
\midrule
Finetuning                           & 24.4 (-10.5) & 26.5 (-7.7) & 17.6 (-16.8) & 28.9 (0.0) & 24.4 \\
Freezing                             & 34.9 (0.0) & 29.4 (0.0) & 30.1 (0.0) & 30.5 (0.0) & 31.2 \\
Joint                                & 45.7 (+10.8) & 40.3 (+4.8) & 43.2 (-1.1) & 40.2 (0.0) & 42.4 \\
\midrule
EWC~\cite{kirkpatrick2017overcoming} & 24.2 (-10.7) & 28.2 (-2.0) & 25.2  (-5.6) & 34.3 (0.0) & 28.0 \\
HAT~\cite{serra2018overcoming}       & 25.7 (-1.0) & 25.5 (-2.7) & 30.1 (-2.1) & 34.4 (0.0) & 28.9 \\
PackNet~\cite{mallya2018packnet}     & 32.5 (0.0) & 32.9 (0.0) & 36.7 (0.0) & 34.3 (0.0) & 34.1 \\
TFM w/o FN                              & 35.3 (0.0) & 38.3 (0.0) & 39.2 (0.0) & 38.0 (0.0) & 37.7 \\
TFM                                  & 35.3 (0.0) & 37.2 (0.0) & 42.0 (0.0) & 37.2 (0.0) & \textbf{38.0} \\
\bottomrule
\end{tabular}}
\end{table}
}
\newcommand{\imagenettable}[1]{
\begin{table*}
\setlength\tabcolsep{4pt}
\caption{#1}
\label{tab:exp_imagenet}
\centering
\vspace{0.1cm}
\resizebox{\textwidth}{!}{
\begin{tabular}{cccccccccccc}
\toprule
\multicolumn{12}{c}{\textbf{ImageNet - classes randomly split}} \\
\toprule
\multirow{ 2}{*}{Approach} & Task 1 & Task 2 & Task 3 & Task 4 & Task 5 & Task 6 & Task 7 & Task 8 & Task 9 & Task 10 & Avg. \\
& (1-100) & (101-200) & (201-300) & (301-400) & (401-500) & (501-600) & (601-700) & (701-800) & (801-900) & (901-1000) & all \\
\midrule
Finetuning & 25.8 (-43.0) & 32.2 (-36.2) & 31.4 (-35.3) & 37.8 (-27.7) & 39.1 (-27.7) & 43.7 (-25.7) & 46.0 (-22.8) & 50.0 (-16.5) & 53.4 (-12.1) & 63.7 (0.0) & 42.3 \\
Freezing & 68.8 (0.0) & 53.5 (0.0) & 52.0 (0.0) & 51.2 (0.0) & 51.3 (0.0) & 53.9 (0.0) & 52.2 (0.0) & 53.9 (0.0) & 51.7 (0.0) & 51.2 (0.0) & 54.0 \\
\midrule
LwF~\cite{li2017learning} & 27.6 (-41.2) & 37.2 (-19.9) & 42.0 (-22.6) & 44.4 (-20.9) & 50.5 (-14.1) & 56.6 (-11.3) & 57.9 (-9.1) & 61.2 (-5.0) & 62.0 (-1.3) & 62.7 (0.0) & 50.2 \\
IMM-mode~\cite{lee2017overcoming} & 68.5 (-0.3) & 53.6 (0.0) & 52.1 (0.0) & 51.7 (-0.1) & 52.5 (+0.3) & 55.5 (+0.2) & 54.7 (+0.1) & 53.5 (0.0) & 54.2 (+0.1) & 51.8 (0.0) & 54.8 \\
EWC~\cite{kirkpatrick2017overcoming} & 21.8 (-47.0) & 26.5 (-41.7) & 29.5 (-36.5) & 32.9 (-32.6) & 35.6 (-30.9) & 40.4 (-28.1) & 40.0 (-26.2) & 44.7 (-20.7) & 47.8 (-16.2) & 61.1 (0.0) & 38.0 \\
PackNet~\cite{mallya2018packnet} & 67.5 (0.0) & 65.8 (0.0) & 62.2 (0.0) & 58.4 (0.0) & 58.6 (0.0) & 58.7 (0.0) & 56.0 (0.0) & 56.5 (0.0) & 54.1 (0.0) & 53.6 (0.0) & 59.1 \\
TFM (Ours) & 63.6 (0.0) & 62.2 (0.0) & 60.1 (0.0) & 61.6 (0.0) & 62.6 (0.0) & 64.5 (0.0) & 64.0 (0.0) & 63.7 (0.0) & 63.0 (0.0) & 59.9 (0.0) & \textbf{62.5} \\
\bottomrule
\end{tabular}}
\end{table*}}

\newcommand{\ternarythreetasks}{
\begin{figure}[t]
    \centering
    \includegraphics[width=0.95\linewidth]{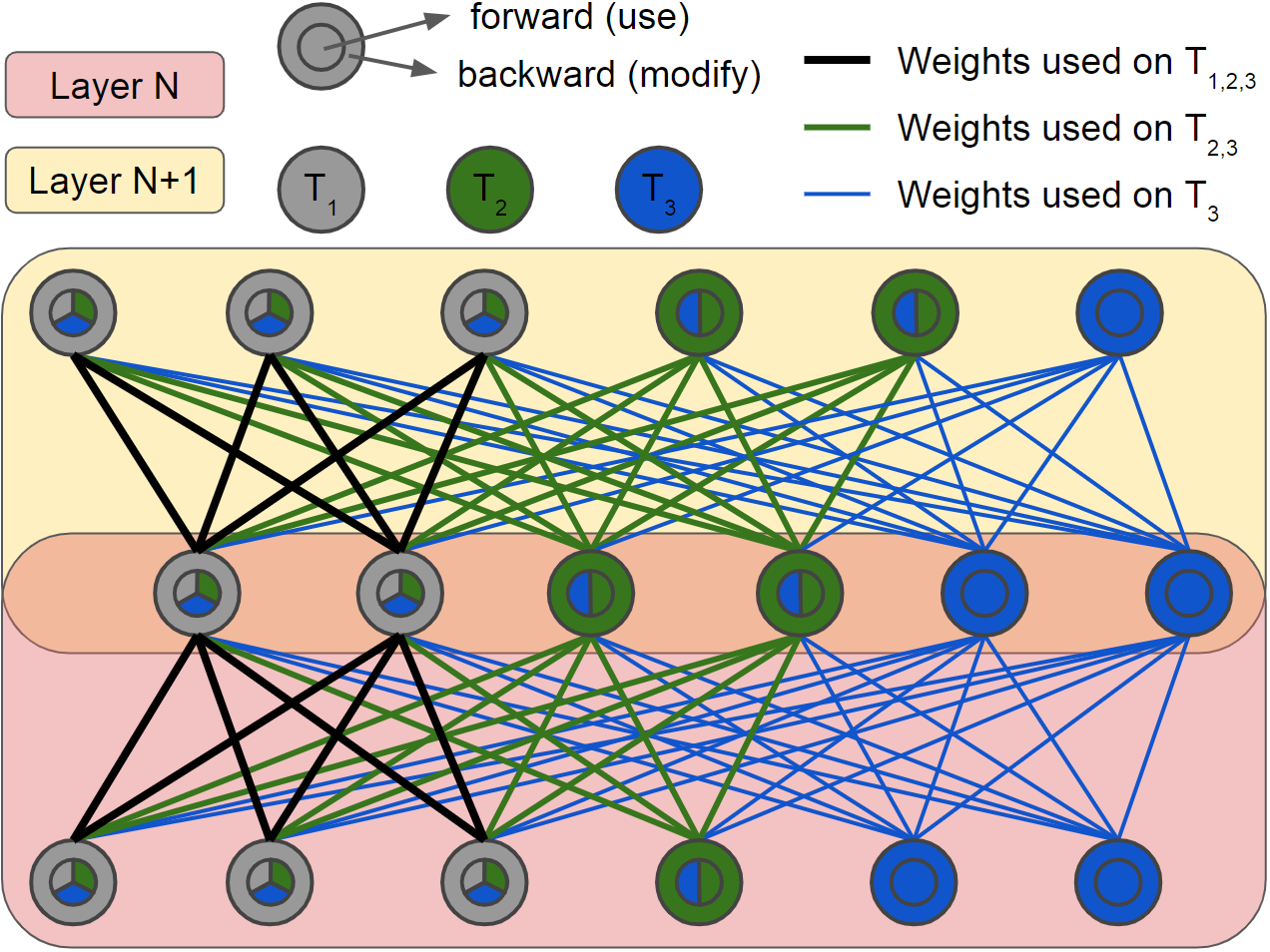}
    \caption{Network growth with ternary feature masks over three tasks.}
    \label{fig:3tasks}
\end{figure}
}
\newcommand{\tinygroupsplits}[1]{
\begin{table}[ht!]
\setlength\tabcolsep{2pt}
\caption{Semantically similar splits for tiny ImageNet.}
\label{tab:tiny_group_split_classes}
\centering
\resizebox{\linewidth}{!}{
\begin{tabular}{ccp{#1}}
\toprule
\textbf{Task} & \textbf{Semantic group} & \textbf{Classes} \\
\midrule
1 & Animals & \multirow{ 4}{#1}{scorpion, black widow, tarantula, spider web, centipede, trilobite, grasshopper, stick insect, cockroach, mantis, ladybug, dragonfly, monarch butterfly, sulphur butterfly, fly, bee, goose, black stork, king penguin, albatross.} \\
 & (flying \& insects) & \\
 & & \\
 & & \\
\midrule
2 & Artifacts & \multirow{ 4}{#1}{abacus, binoculars, candle, chain, chest, dumbbell, hourglass, lampshade, magnetic compass, pill bottle, computer keyboard, acorn, plunger, syringe, teddy bear, torch, comic book, remote control, umbrella, nail.} \\
 & (smaller) & \\
 & & \\
 & & \\
\midrule
3 & Music, Sport & \multirow{ 4}{#1}{basketball, punching bag, rugby ball, scoreboard, stopwatch, volleyball, CD player, drumstick, iPod, oboe, organ, refrigerator, cask, plate, wooden spoon, teapot, frying pan, beaker, bucket, dining table.} \\
 & and Kitchen & \\
 & & \\
 & & \\
\midrule
4 & Animals & \multirow{ 5}{#1}{brown bear, red panda, koala, pig, ox, bison, bighorn sheep, gazelle, dromedary, African elephant, orangutan, chimpanzee, baboon, cougar, lion, European fire salamander, bullfrog, tailed frog, American alligator, boa constrictor.} \\
 & (land) & \\
 & & \\
 & & \\
 & & \\
\midrule
5 & Artifacts & \multirow{ 4}{#1}{altar, maypole, bannister, flagpole, fountain, parking meter, pay-phone, pole, cash machine, birdhouse, reel, bathtub, rocking chair, potter's wheel, sewing machine, space heater, turnstile, memorial tablet, desk, vestment.} \\
 & (bigger) & \\
 & & \\
 & & \\
\midrule
6 & Food & \multirow{ 4}{#1}{water jug, wok, guacamole, ice cream, lollipop, pretzel, mashed potato, cauliflower, bell pepper, mushroom, orange, lemon, banana, pomegranate, meat loaf, pizza, potpie, espresso, soda bottle, beer bottle.} \\
 & & \\
 & & \\
 & & \\
\midrule
7 & Animals & \multirow{ 5}{#1}{dugong, goldfish, jellyfish, brain coral, American lobster, spiny lobster, sea slug, sea cucumber, guinea pig, snail, slug, poodle, Chihuahua, Yorkshire terrier, golden retriever, Labrador retriever, German shepherd, tabby cat, Persian cat, Egyptian cat.} \\
 & (water \& pets) & \\
 & & \\
 & & \\
 & & \\
\midrule
8 & Clothes and & \multirow{ 4}{#1}{academic gown, poncho, apron, backpack, bikini, bow tie, fur coat, gasmask, kimono, sock, military uniform, miniskirt, neck brace, Christmas stocking, sombrero, sunglasses, cardigan, snorkel, sandal, swimming trunks.} \\
 & wearables & \\
 & & \\
 & & \\
\midrule
9 & Transport & \multirow{ 4}{#1}{bullet train, station wagon, freight car, go-kart, rickshaw, lifeboat, limousine, moving van, police van, school bus, convertible, crane, trolleybus, sports car, tractor, gondola, broom, cannon, lawn mower, missile.} \\
 & & \\
 & & \\
 & & \\
\midrule
10 & Buildings & \multirow{ 5}{#1}{barbershop, barn, lighthouse, butcher shop, candy store, water tower, triumphal arch, suspension bridge, steel arch bridge, viaduct, thatched roof, cliff dwelling, dam, obelisk, picket fence, cliff, coral reef, lakeside, seacoast, alp.} \\
 & and scenes & \\
 & & \\
 & & \\
 & & \\
\bottomrule
\end{tabular}}
\end{table}}
\newcommand{\multipledatasets}[2]{
\begin{figure}[t]
    \centering
    \includegraphics[trim=#1, clip, width=\linewidth]{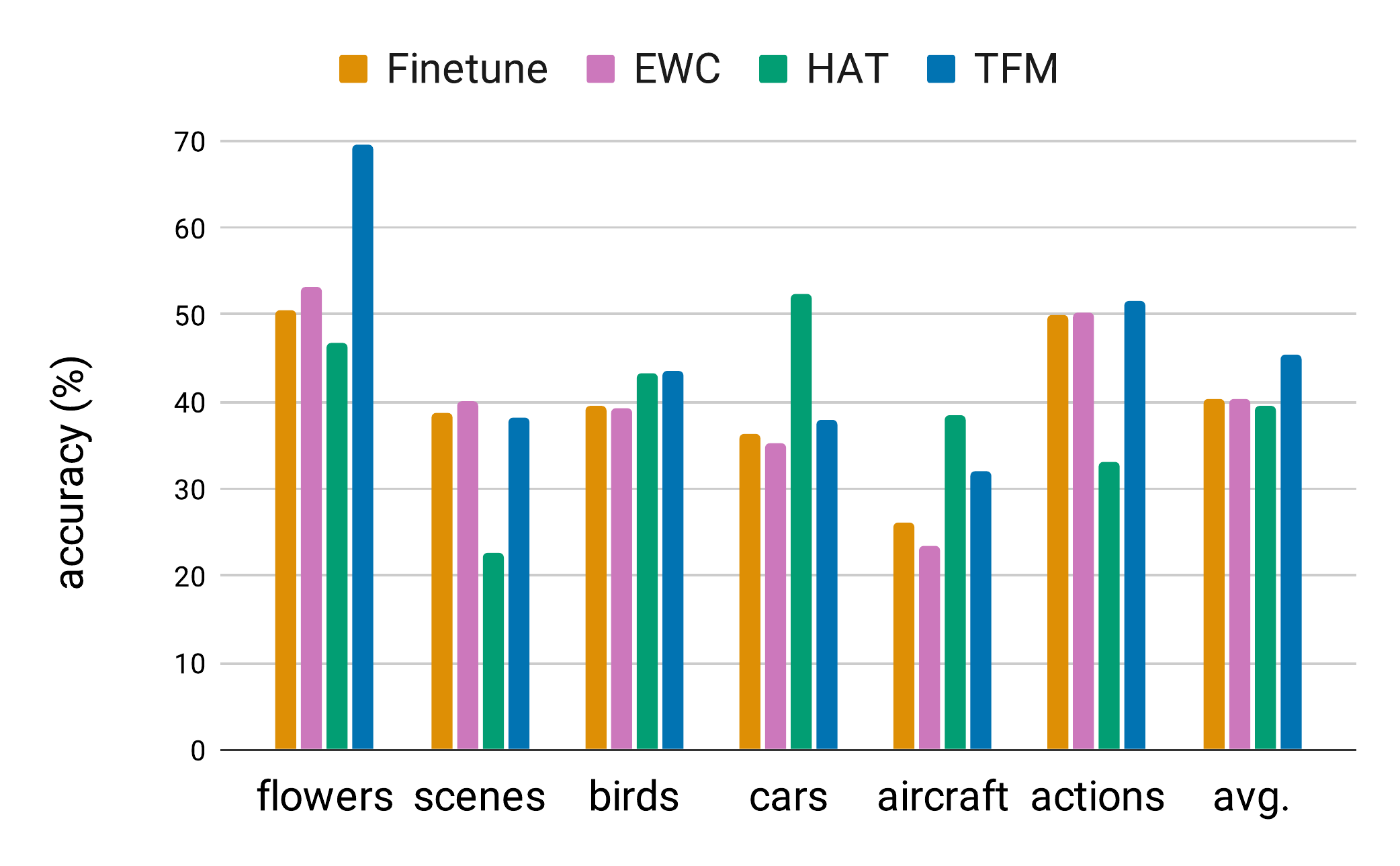}
    \caption{#2}
    \label{fig:multi-dataset}
\end{figure}
}
\newcommand{\tinytable}[1]{
\begin{table}
\caption{#1}
\label{tab:tiny_imagenet_all}
\vspace{0.1cm}
\centering
\resizebox{0.8\linewidth}{!}{
\begin{tabular}{cccc}
\toprule
\multicolumn{4}{c}{\textbf{Tiny ImageNet \,--\, avg. acc. after 10 tasks}} \\
\toprule
\multirow{ 2}{*}{Approach} & Random & Semantic & Larger \\
& split & split & 1st task \\
\midrule
Finetuning & 47.4 & 28.0 & 57.2 \\
Freezing   & 42.7 & 32.8 & 69.9 \\
\midrule
LfL~\cite{jung2016less}              & 47.7 & 27.8 & 60.0 \\
LwF~\cite{li2017learning}            & 56.4 & 37.8 & 61.1 \\
IMM-mode~\cite{lee2017overcoming}    & 48.3 & 33.9 & 62.0 \\
EWC~\cite{kirkpatrick2017overcoming} & 47.8 & 27.8 & 56.6 \\
HAT~\cite{serra2018overcoming}       & 54.2 & 44.0 & 66.5 \\
PackNet~\cite{mallya2018packnet}     & \textbf{56.4} & 45.2 & 70.8 \\
\midrule
TFM w/o FN (Ours) & 54.9 & 44.3 & 72.4 \\
TFM (Ours)        & 56.0 & \textbf{45.3} & \textbf{73.3} \\
\bottomrule
\end{tabular}}
\end{table}}
\newcommand{\tinyrandomtable}[1]{
\begin{table*}
\setlength\tabcolsep{4pt}
\caption{#1}
\label{tab:tiny_imagenet_random}
\vspace{0.2cm}
\centering
\resizebox{\textwidth}{!}{
\begin{tabular}{cccccccccccc}
\toprule
\multicolumn{12}{c}{\textbf{Tiny ImageNet - classes randomly split}} \\
\toprule
\multirow{ 2}{*}{Approach} & Task 1 & Task 2 & Task 3 & Task 4 & Task 5 & Task 6 & Task 7 & Task 8 & Task 9 & Task 10 & Avg. \\
& (1-20) & (21-40) & (41-60) & (61-80) & 81-100) & (101-120) & (121-140) & (141-160) & (161-180) & (181-200) & all \\
\midrule
Finetuning & 38.1 (-13.6) & 36.0 (-13.7) & 43.2 (-16.0) & 44.1 (-18.6) & 45.5 (-12.6) & 54.5 (-13.6) & 50.3 (-15.7) & 50.5 (-13.4) & 51.0 (-13.1) & 61.2 (0.0) & 47.4 \\
Freezing & 51.7 (0.0) & 36.4 (0.0) & 39.5 (0.0) & 41.7 (0.0) & 42.9 (0.0) & 46.2 (0.0) & 45.7 (0.0) & 41.1 (0.0) & 41.2 (0.0) & 40.9 (0.0) & 42.7 \\
Joint & 58.6 (+6.9) & 53.9 (+8.3) & 59.1 (+3.7) & 61.8 (+7.9) & 57.7 (+2.9) & 66.0 (+2.6) & 64.0 (+3.1) & 60.2 (+5.9) & 57.9 (+1.0) & 53.8 (0.0) & 59.3 \\
\midrule
LfL~\cite{jung2016less} & 32.4 (-18.9) & 35.4 (-17.0) & 43.4 (-15.7) & 44.1 (-20.2) & 45.0 (-15.0) & 55.9 (-14.5) & 49.4 (-16.1) & 51.1 (-12.4) & 58.6 (-8.0) & 61.4 (0.0) & 47.7 \\
LwF~\cite{li2017learning} & 45.1 (-6.6) & 45.5 (-2.2) & 53.5 (-4.6) & 57.6 (-2.6) & 56.2 (0.0) & 65.7 (+0.4) & 63.5 (-0.3) & 58.4 (-1.9) & 59.6 (-0.3) & 58.5 (0.0) & 56.4 \\
IMM-mode~\cite{lee2017overcoming} & 50.6 (-1.1) & 38.5 (+0.3) & 44.7 (-0.1) & 49.2 (+0.3) & 47.5 (+1.1) & 51.9 (-1.4) & 53.7 (-0.6) & 47.7 (-0.4) & 50.0 (-2.2) & 48.7 (0.0) & 48.3 \\
EWC~\cite{kirkpatrick2017overcoming} & 33.9 (-17.4) & 35.4 (-14.4) & 43.6 (-15.4) & 46.7 (-15.9) & 49.5 (-9.1) & 52.5 (-15.8) & 47.8 (-20.0) & 50.2 (-13.8) & 56.6 (-9.9) & 61.4 (0.0) & 47.8 \\
HAT~\cite{serra2018overcoming} & 46.8 (-0.2) & 49.1 (+0.8) & 55.8 (+0.2) & 58.0 (-0.2) & 53.7 (+0.3) & 61.0 (+0.1) & 58.7 (0.0) & 54.0 (-0.1) & 54.6 (-0.1) & 50.3 (0.0) & 54.2 \\
PackNet~\cite{mallya2018packnet} & 52.5 (0.0) & 49.7 (0.0) & 56.5 (0.0) & 59.8 (0.0) & 55.0 (0.0) & 64.7 (0.0) & 61.7 (0.0) & 55.9 (0.0) & 55.2 (0.0) & 52.5 (0.0) & \textbf{56.4} \\
\midrule
TFM w/o FN (Ours) & 49.6 (0.0) & 47.2 (0.0) & 54.8 (0.0) & 58.2 (0.0) & 55.0 (0.0) & 64.0 (0.0) & 59.3 (0.0) & 53.6 (0.0) & 55.5 (0.0) & 51.9 (0.0) & 54.9 \\
TFM (Ours) & 48.2 (0.0) & 47.7 (0.0) & 56.7 (0.0) & 58.2 (0.0) & 54.8 (0.0) & 62.2 (0.0) & 61.5 (0.0) & 57.3 (0.0) & 58.5 (0.0) & 54.8 (0.0) & 56.0 \\
\bottomrule
\end{tabular}}
\end{table*}}
\newcommand{\tinygroupstable}[1]{
\begin{table*}
\setlength\tabcolsep{4pt}
\caption{#1}
\label{tab:tiny_imagenet_groups}
\vspace{0.2cm}
\centering
\resizebox{\textwidth}{!}{
\begin{tabular}{cccccccccccc}
\toprule
\multicolumn{12}{c}{\textbf{Tiny ImageNet - classes semantically split}} \\
\toprule
\multirow{ 2}{*}{Approach} & Task 1 & Task 2 & Task 3 & Task 4 & Task 5 & Task 6 & Task 7 & Task 8 & Task 9 & Task 10 & Avg. \\
& fly anim. & small artif. & hobbies & land anim. & big artif. & food & pets/aquatic & wearables & transport & scenes & all \\
\midrule
Finetuning & 17.1 (-34.2) & 19.7 (-17.7) & 20.9 (-24.5) & 16.7 (-30.5) & 20.8 (-28.7) & 29.2 (-22.0) & 30.7 (-21.0) & 25.2 (-17.8) & 40.2 (-18.9) & 59.9 (0.0) & 28.0 \\
Freezing & 51.3 (0.0) & 28.5 (0.0) & 27.2 (0.0) & 29.6 (0.0) & 29.0 (0.0) & 35.0 (0.0) & 31.7 (0.0) & 23.9 (0.0) & 37.0 (0.0) & 34.7 (0.0) & 32.8 \\
Joint & 55.0 (+3.7) & 41.9 (+7.9) & 46.2 (+6.3) & 44.9 (+4.9) & 44.7 (+7.1) & 49.0 (+4.8) & 46.6 (+4.9) & 36.4 (+4.7) & 51.2 (+5.7) & 51.1 (0.0) & 46.7 \\
\midrule
LfL~\cite{jung2016less} & 17.2 (-34.1) & 18.4 (-21.0) & 21.5 (-24.0) & 18.7 (-30.5) & 20.2 (-28.6) & 27.4 (-23.9) & 28.4 (-22.3) & 26.0 (-18.4) & 41.2 (-17.6) & 59.1 (0.0) & 27.8 \\
LwF~\cite{li2017learning} & 34.0 (-13.9) & 18.4 (-14.5) & 32.6 (-0.8) & 36.5 (-5.6) & 40.1 (-0.5) & 43.1 (-2.5) & 41.8 (-1.3) & 32.7 (-1.1) & 50.3 (-0.5) & 48.1 (0.0) & 37.8 \\
IMM-mode~\cite{lee2017overcoming} & 42.3 (-9.0) & 28.8 (+0.1) & 26.5 (-3.1) & 30.7 (-3.6) & 32.5 (-3.1) & 28.8 (-13.0) & 35.4 (-6.2) & 27.3 (-3.7) & 43.6 (-4.9) & 42.7 (0.0) & 33.9 \\
EWC~\cite{kirkpatrick2017overcoming} & 20.2 (-31.1) & 18.5 (-19.3) & 20.2 (-26.3) & 20.9 (-28.9) & 24.7 (-22.8) & 25.5 (-27.5) & 28.7 (-23.4) & 23.0 (-19.6) & 39.8 (-20.2) & 56.8 (0.0) & 27.8 \\
HAT~\cite{serra2018overcoming} & 44.6 (+0.4) & 34.8 (+0.2) & 40.8 (-0.1) & 45.4 (+0.4) & 40.8 (-2.5) & 49.8 (0.0) & 44.9 (-0.2) & 33.1 (-1.7) & 51.9 (0.0) & 53.8 (0.0) & 44.0 \\
PackNet~\cite{mallya2018packnet} & 47.0 (0.0) & 35.7 (0.0) & 42.7 (0.0) & 48.6 (0.0) & 45.8 (0.0) & 48.1 (0.0) & 45.9 (0.0) & 38.3 (0.0) & 51.2 (0.0) & 49.1 (0.0) & 45.2 \\
\midrule
TFM w/o FN (Ours) & 46.4 (0.0) & 34.7 (0.0) & 38.8 (0.0) & 44.1 (0.0) & 42.0 (0.0) & 48.3 (0.0) & 46.5 (0.0) & 35.7 (0.0) & 52.0 (0.0) & 54.8 (0.0) & 44.3 \\
TFM (Ours) & 46.4 (0.0) & 37.2 (0.0) & 40.4 (0.0) & 44.1 (0.0) & 44.2 (0.0) & 48.2 (0.0) & 46.4 (0.0) & 37.5 (0.0) & 53.5 (0.0) & 54.7 (0.0) & \textbf{45.3} \\
\bottomrule
\end{tabular}}
\end{table*}}
\newcommand{\tinylargetask}[1]{
\begin{table*}
\setlength\tabcolsep{4pt}
\caption{#1}
\label{tab:tiny_large_task}
\vspace{0.2cm}
\centering
\resizebox{\textwidth}{!}{
\begin{tabular}{cccccccccccc}
\toprule
\multicolumn{12}{c}{\textbf{Tiny ImageNet - larger first task}} \\
\toprule
\multirow{ 2}{*}{Approach} & Task 1 & Task 2 & Task 3 & Task 4 & Task 5 & Task 6 & Task 7 & Task 8 & Task 9 & Task 10 & Avg. \\
& (1-110) & (111-120) & (121-130) & (131-140) & (141-150) & (151-160) & (161-170) & (171-180) & (181-190) & (191-200) & (111+) \\
\midrule
Finetuning & 18.4 (-33.2) & 39.6 (-34.4) & 56.6 (-21.0) & 58.0 (-23.8) & 44.6 (-32.8) & 63.0 (-21.8) & 51.0 (-27.2) & 51.4 (-19.0) & 70.4 (-14.0) & 80.0 (0.0) & 57.2 \\
Freezing & 51.6 (0.0) & 68.6 (0.0) & 70.2 (0.0) & 77.9 (0.0) & 68.1 (0.0) & 78.8 (0.0) & 72.9 (0.0) & 64.2 (0.0) & 76.7 (0.0) & 70.2 (0.0) & 69.9 \\
\midrule
LfL~\cite{jung2016less} & 16.7 (-33.9) & 58.3 (-6.6) & 59.5 (-2.9) & 64.6 (-2.2) & 58.2 (-1.4) & 64.7 (-0.6) & 63.4 (+0.5) & 54.4 (+0.1) & 61.0 (0.0) & 56.5 (0.0) & 60.0 \\
LwF~\cite{li2017learning} & 10.8 (-40.0) & 29.5 (-43.5) & 44.6 (-30.4) & 61.2 (-21.2) & 55.5 (-15.1) & 73.3 (-8.8) & 71.7 (-3.5) & 62.3 (-1.8) & 77.8 (-2.2) & 74.3 (0.0) & 61.1 \\
IMM-mode~\cite{lee2017overcoming} & 26.1 (-26.3) & 50.4 (-13.8) & 59.5 (-19.3) & 61.6 (22.8) & 54.0 (-21.8) & 63.2 (-22.5) & 58.2 (-19.9) & 56.0 (-13.9) & 75.0 (-6.7) & 79.9 (0.0) & 62.0 \\
EWC~\cite{kirkpatrick2017overcoming} & 51.8 (-0.7) & 24.8 (-0.1) & 43.3 (-1.2) & 61.8 (-0.5) & 55.5 (-0.3) & 70.8 (-0.7) & 67.6 (-0.1) & 53.8 (-0.6) & 70.1 (-1.2) & 61.7 (0.0) & 56.6 \\
HAT~\cite{serra2018overcoming} & 46.1 (0.0) & 60.1 (-0.1) & 68.2 (+0.2) & 73.2 (+0.1) & 63.2 (+0.1) & 76.2 (+0.1) & 67.4 (0.0) & 58.1 (-0.1) & 73.2 (0.0) & 59.1 (0.0) & 66.5 \\
PackNet~\cite{mallya2018packnet} & 47.6 (0.0) & 74.0 (0.0) & 74.2 (0.0) & 79.0 (0.0) & 65.2 (0.0) & 76.2 (0.0) & 69.4 (0.0) & 61.4 (0.0) & 73.4 (0.0) & 64.8 (0.0) & 70.8 \\
\midrule
TFM w/o FN (Ours) & 49.6 (0.0) & 69.8 (0.0) & 71.1 (0.0) & 79.8 (0.0) & 68.4 (0.0) & 78.4 (0.0) & 72.9 (0.0) & 64.8 (0.0) & 75.9 (0.0) & 70.2 (0.0) & 72.4 \\
TFM (Ours) & 49.9 (0.0) & 70.4 (0.0) & 71.4 (0.0) & 80.8 (0.0) & 70.5 (0.0) & 79.4 (0.0) & 73.9 (0.0) & 64.4 (0.0) & 76.5 (0.0) & 72.1 (0.0) & \textbf{73.3} \\
\bottomrule
\end{tabular}}
\end{table*}}

\newcommand{\commentgrey}[1]{{\color{gray}{\textit{#1}}}}
\newcommand{\minisection}[1]{\vspace{0.04in} \noindent {\bf #1}\ \ }

\fancypagestyle{mystyle}{%
    \fancyhead[C]{\color{gray}{\sffamily \small{{\shortstack This is the author’s version of an article to appear in IEEE Conference on \\Computer Vision and Pattern Recognition Workshop (CVPR-W) on Continual Learning in Computer Vision (CLVision) 2021.}}}
    }
}%
\begin{document}

\title{Ternary Feature Masks: zero-forgetting for task-incremental learning}

\author{Marc Masana\\
Computer Vision Center\\
Barcelona, Spain\\
{\tt\small mmasana@cvc.uab.es}
\and
Tinne Tuytelaars\\
ESAT-PSI\\
KU Leuven, Belgium\\
{\tt\small tinne.tuytelaars@esat.kuleuven.be}
\and
Joost van de Weijer\\
Computer Vision Center\\
Barcelona, Spain\\
{\tt\small joost@cvc.uab.es}
}

\maketitle
\thispagestyle{mystyle}

\begin{abstract}
We propose an approach without any forgetting to continual learning for the task-aware regime, where at inference the task-label is known. By using ternary masks we can upgrade a model to new tasks, reusing knowledge from previous tasks while not forgetting anything about them. Using masks prevents both catastrophic forgetting and backward transfer. We argue -- and show experimentally -- that avoiding the former largely compensates for the lack of the latter, which is rarely observed in practice. In contrast to earlier works, our masks are applied to the features (activations) of each layer instead of the weights. This considerably reduces the number of mask parameters for each new task; with more than three orders of magnitude for most networks. The encoding of the ternary masks into two bits per feature creates very little overhead to the network, avoiding scalability issues. To allow already learned features to adapt to the current task without changing the behavior of these features for previous tasks, we introduce task-specific feature normalization. Extensive experiments on several finegrained datasets and ImageNet show that our method outperforms current state-of-the-art while reducing memory overhead in comparison to weight-based approaches.
\end{abstract}
\section{Introduction}
\label{sec:intro}
Fine-tuning has been established as the most common method to use when learning a new task on top of an already learned one. This works well if you no longer require the system to perform the previous task. However, in many real-world situations one is interested in learning consecutive tasks, all of which, in the end, the system should be able to perform all. This is the setting studied in lifelong learning, also referred to as sequential, incremental or continual learning. In this setting, the popular approach of fine-tuning suffers from catastrophic forgetting~\cite{french1999catastrophic, goodfellow2014empirical, li2020baseline, mccloskey1989catastrophic, mermillod2013stability}: all network capabilities are used for learning the new task, which leads to forgetting of the previous ones.

A popular strategy to avoid this is to use importance-weight loss proxies or regularizers~\cite{aljundi2018memory, kirkpatrick2017overcoming, lee2017overcoming}. These approaches compute an importance score for each of the weight parameters of the model based on previous tasks and use this to decide which weights can be modified for the current task. A drawback of these methods is that they store an extra variable (the importance score) for each weight. This leads to an overhead of a float per weight parameter, i.e. double the number of parameters which have to be stored. Other methods work with a binary mask to select part of the model for each task~\cite{mallya2018piggyback, mallya2018packnet}. This leads to an overhead of one bit per task per weight parameter. Finally, some methods directly make a copy of the network~\cite{rusu2016progressive} or rely on the storage of exemplars~\cite{chaudhry2018efficient, lopez2017gradient, rebuffi2016icarl, riemer2018learning}, which again increases memory consumption and renders these methods unsuitable when privacy requirements forbid storing of data.

In this paper, we advocate computing a mask at the level of the features 
instead of at the level of the weights. We need the mask to be ternary, i.e.~adding a third state allowing features to be used during the forward pass while being masked during the backward pass. This allows to reuse representations from previous tasks without introducing forgetting and drastically reduces the number of extra parameters that need to be stored. As an example, the popular AlexNet architecture~\cite{krizhevsky2012imagenet} has around 60m weights, while having less than 10k features. One earlier method that builds on this idea is HAT~\cite{serra2018overcoming}, which stores an attention value for each feature for each task. Recently, SSL~\cite{aljundi2018selfless} also brings attention to the activation neurons by promoting sparsity with losses inspired by lateral inhibition in the mammalian brain. Those two recent works stress the importance of focusing on features instead of weights, not only because of the reduction in memory overhead, but also because they allow for better performance and less forgetting. However, both methods still allow some forgetting as new tasks are learned.

Over time, the forgetting typically increases with the number of tasks~\cite{aljundi2018memory, kirkpatrick2017overcoming, lee2017overcoming, lopez2017gradient, rebuffi2016icarl, serra2018overcoming}. However, for many practical systems, it is undesirable if the accuracy of the system deteriorates over time, while the system learns new tasks. Moreover, under these settings, the user typically has no control on the amount of forgetting, i.e. there are no guarantees on the performance of the system after new tasks have been added. Even worse: to the user, it is unknown how much the system has actually forgotten or how well the system still performs on older tasks.

For these reasons, some works have studied continual learning systems without any forgetting at all. Currently, apart from the methods that make copies of the network for each task~\cite{rusu2016progressive}, mask-based approaches are the only ones that guarantee no-forgetting~\cite{mallya2018piggyback, mallya2018packnet}. Indeed, all methods that allow backward propagation into the parameters of previously learned tasks have no control on the amount of forgetting. Not updating the weights used for previous tasks using a binary mask prevents any forgetting. In the case of recent approaches PackNet~\cite{mallya2018packnet} and PiggyBack~\cite{mallya2018piggyback}, this is enforced by binary-masking weights or learning masks that will be binarized after the task is learned. However, both these non-forgetting methods mask weights and therefore have a larger memory overhead than methods based on feature-masks. Another drawback of~\cite{mallya2018piggyback} is that it requires a backbone network as a starting point.

We propose Ternary Feature Masks (TFM), a method for continual learning which does not suffer from any forgetting. Due to the nature of our proposed mask-based approach, we will only focus on evaluation on task-aware experimental setups. Instead of applying masks to all weights in the network, we propose to move the masks to the feature level, thereby significantly reducing memory overhead. Our initial method requires only a 2-bit mask value for each activation for each task. In addition, we introduce a task-dependent feature normalization (FN). This allows to adjust previously learned features to be of more optimal use for later tasks, without changing the performance or weights assigned to previous tasks. This introduces a further memory overhead of storing two floats more per activation per task. Nevertheless, this method still has a significantly lower memory overhead than any method which stores additional parameters per weight~\cite{aljundi2018memory, kirkpatrick2017overcoming, lee2017overcoming, liu2018rotate, mallya2018piggyback, mallya2018packnet, zenke2017continual}.

Mask-based approaches have shown to be better at overcoming catastrophic forgetting on task-incremental learning (task-IL)~\cite{delange2021continual,mallya2018packnet, serra2018overcoming}. However, unlike most distillation and model-based approaches, they make use of some overhead memory during inference. In Fig.~\ref{fig:overhead} we visualize the absolute memory overhead used by some approaches on an ImageNet scenario with 10 tasks. Considering that the network used is around 220Mb, we can observe that the approaches that focus on using feature-masks (HAT, TFM) have a negligible overhead in comparison to the weight-masked approaches (PackNet). It should also be noticed that mask-based approaches keep the same overhead during training, while distillation and model-based approaches usually duplicate the network size at least. In conclusion, our method has similar memory usage as HAT, however, we outperform this method on all proposed experiments, and our method is significantly more memory efficient than PackNet whose performance we either match or outperform.

\imagenetoverhead{Log scale overhead growth for ImageNet on AlexNet. Best viewed in color.}

\section{Related work}
\label{sec:soa}
Continual learning in the proposed task-IL setup has been addressed in multiple prior works~\cite{delange2021continual, lesort2020continual, parisi2019continual, pfulb2019comprehensive}. A large part of the approaches use regularization-based techniques to reduce catastrophic forgetting without having to store raw input. They can be divided into two main families: distillation approaches and model-based approaches.

Distillation approaches use teacher-student setups that aim at preserving the output of the teacher model on the new data distribution. LwF proposes to use the knowledge distillation loss~\cite{li2017learning} to preserve the performance of previous tasks. However, if the data distribution of the new task is very different from the previous tasks, performance drops drastically~\cite{aljundi2016expert}. In order to solve that, iCaRL~\cite{rebuffi2016icarl} stores a subset of each tasks' data as exemplars; while EBLL~\cite{rannen2017encoder} solves the issue by learning undercomplete autoencoders for each task. LFL is also similar to LwF, preserving the previous tasks' performance by penalizing changes on the shared representation~\cite{jung2016less}. Expert Gate~\cite{aljundi2016expert} learns a model for each task and an autoencoder gate which will choose the model to be used. Recently,~\cite{lee2019overcoming, zhang2020class} propose to learn new classes separately and then learn a final model with multiple distillation and extra unlabelled data. However, most of these methods need a pre-processing step before each task. Furthermore, a main issue is also the scalability when learning many tasks, since the described methods have to store data, autoencoders, or larger models for each new task.

Most model-based approaches, when learning a new task, apply a smooth penalty for changing weights, proportional to their importance for previous tasks~\cite{aljundi2018memory, kirkpatrick2017overcoming, lee2017overcoming, liu2018rotate, zenke2017continual}. One of the main issues is they might over or under-estimate the importance of those weights. The main difference among those methods is how that importance is calculated. In EWC an approximation of the diagonal of the Fisher Information Matrix (FIM) is used~\cite{kirkpatrick2017overcoming}. R-EWC proposes a rotation of the weight space to get a better approximation of the FIM~\cite{liu2018rotate}. In IMM the moments of the posterior distribution are matched incrementally~\cite{lee2017overcoming}. SI computes the importance weights in an online fashion by storing how much the loss would change for each parameter over the training~\cite{zenke2017continual}. MAS computes the weight importance in an online unsupervised way, connecting their approach with Hebbian learning~\cite{aljundi2018memory}.

Some more works use other underlying methods. PNN add lateral connections at each layer of the network to a duplicate of that layer~\cite{rusu2016progressive}. Then, the new column learns the new task while the old one keeps the weights fixed, meaning that resources are duplicated each time a task is added. This approach leads to zero-forgetting while making the knowledge of previous tasks available during the learning of a new one through distillation. However, as each new task adds a column with the corresponding connections, the overhead scales drastically with the number of tasks. Progress and Compress expands the idea of PNN with the use of EWC but keeping the number of parameters constant~\cite{schwarz2018progress}. They propose a two-component setup with a knowledge base and an active column that follows a similar setup as PNN. Recently, Learn to Grow allows for each layer to reuse existing weights, adapt them or grow the network~\cite{li2019learn}. In the worst case scenario, layers are added which makes the growth linear in the number of tasks. Finally, ACL combines an architecture growth with experience replay to learn task-specific and task-invariant features~\cite{ebrahimi2020adversarial}. However, this comes at a quite large overhead per task, and we have not been able to obtain competitive results outside of their proposed small datasets (i.e.~below Finetuning for Flowers 4 tasks).

Apart from the above mentioned families, some recent works use masks to directly influence or completely remove forgetting. We refer to this family of approaches as mask-based. PathNet uses evolutionary strategies to learn selective routing through the weights~\cite{fernando2017pathnet}. However, it is not end-to-end differentiable and computationally very expensive. PackNet trains with available weights, then prunes the less relevant ones and retrains with a smaller subset of them~\cite{mallya2018packnet}. Those weights are then not available for further learning of new tasks, which quickly reduces the capacity of the network. This results in lower number of parameters being free and performance dropping quickly on longer sequences. Piggyback proposes to use a pretrained network as a backbone and then uses binary masks on the weights to create different sub-networks for each task~\cite{mallya2018piggyback}. Its main drawback is the backbone network itself, which is crucial to being able to learn each task on top of it and cannot have a too different distribution from them. Finally, HAT proposes a hard attention mechanism on the features after each layer~\cite{serra2018overcoming}. The attention embeddings are non-binary and are learned together with each task and conditioned by the attentions of previous ones. This offers plasticity to the embeddings in order to learn them, but also allows the possibility to forget previous tasks during the back-propagation step. A zero-forgetting idea is discussed in the appendices of their manuscript with a note on binary masks, connecting the removal of plasticity to \emph{inhibitory synapses}~\cite{mcculloch1943logical}.

In our approach we take the latter side of that balance, using rigid masks that reduce plasticity but also ensure non-forgetting of previous tasks. Our approach also focuses on a natural expansion of the capacity of the network, which is not addressed in HAT and most of the previous related work. Our approach uses masks on the features of the network to have a better control over which weights can be modified while learning new tasks. At the same time, the mask being ternary allows weights fixed for previous tasks to be used on new tasks without modifying those weights. This masking strategy allows the network to not forget anything from previous tasks and reduce the computational overhead in comparison to masking the weights. Our proposed method is unique in that it combines being expandable, having a low overhead cost and having no forgetting. All other methods have to choose only one of those three characteristics if any.

\section{Learning without any Forgetting}
\label{sec:method}
Here we propose our zero-forgetting method for task-aware incremental learning. As discussed in the introduction, in order to enforce non-forgetting of previous tasks, the use of masks that create rigid states is an efficient way. Works which have addressed this problem have focused on \emph{weight-masks} where an additional parameter is learned for each weight in the network~\cite{mallya2018piggyback, mallya2018packnet}. From a network overhead point of view, we argue that it is, however, better to work with \emph{feature-masks} which learn an additional parameter for each feature in the network. In Table~\ref{table:networks} we compare the number of weights and features in several popular networks. The table clearly shows that the overhead is significantly lower: on average weight-masks are a quadratic factor bigger than feature-masks.

\weightsfeatures{Difference between number of weights and features for different common network architectures (without heads).}

First, we discuss binary masks and how those can easily encode the parts that we want to learn and the parts that we want to fix. Afterwards, we explore what happens when we want to learn more than one task and extending to ternary masks. Finally, we explore the use of feature normalization to allow for less rigid learning of new tasks.

\subsection{Binary feature masks}
Using binary feature masks on neural networks means that the masked neuron will have one of two states (0 or 1). When the masks are directly multiplied by the neuron activations, the corresponding filters will be used or not (same for the backward pass, which will either be applied or not). Then, for each task we have a binary mask with the neurons that can be used. Since we pursue zero-forgetting, those masks will have to be disjoint. In Fig.~\ref{fig:binary_mask} we show an example with two tasks where each of them is only allowed to use different neurons. A large amount of connections are completely unused, making the two sub-networks totally separable from one another.

Consider a fully-connected layer (the theory can easily be extended to convolutional layers). The output of the layer is $y= W x$ where $y\in \mathbb{R}^{p\times1}$, $x \in \mathbb{R}^{q\times1}$ and $W \in \mathbb{R}^{p\times q}$. The binary feature mask for the forward pass is defined as:
\begin{equation}
y=\left(Wx\right)\odot m^{t,l}
\end{equation}
where $m^{t,l}\in \mathbb{R}^{p\times1}$ refers to the mask for task $t$ at layer $l$ and $\odot$ is an element-wise multiplication. Masks from different tasks are forced to select different features (\mbox{$\left(m^{s,l}\right)^{\rm T}\!\cdot\,m^{t,l}\;=\;0$} \medmuskip=0mu $\forall\,s\neq\,t)$).
The backward pass for training task $t$ is defined as:
\begin{equation}
\frac{\partial \mathcal{L}}{\partial W_{ij}} = (m^{t,l}_{i}\ \wedge\ m^{t,l-1}_{j}) x_j\frac{\partial \mathcal{L}}{\partial y_{i}}
\end{equation}
where $\wedge$ is the AND logical operator and there are only non-zero gradients for those weights which join in a feature which is masked for task $t$.

This setup allows the associated weights to an activation to be either used-and-learnable, or neither. If used, they will contribute forward to the next layers (which is good, as it promotes forward transfer, i.e. sharing of knowledge from previous tasks). Yet at the same time this also implies that it will be possible to modify them (which is bad, as it introduces catastrophic forgetting on previous tasks).
With only binary masks, you cannot have one without the other. Alternatively, one could also define two separate binary states: ``used" and `learnable". This has been used for a long time in deep learning by freezing weights~\cite{oquab2014learning}. Freezing weights is a mask-based way of switching on and off the learning of a layer. In this case, in both states the layer would contribute to the outcome of the network, but the update of the weights would only be done on those layers that are not masked. Here, we further explore this idea. We advocate that, in a sequential setup where the capacity of the network might increase when learning new tasks, the best way to mask the neurons is by having three states: ``used", ``learnable" and ``unused". This can be achieved by using ternary masks on the neurons.

\subsection{Ternary feature masks (TFM)}
\label{sec:ternary_feat_masks}
Being able to use the connections between the neurons of the previous tasks and the neurons of the newly added task is important to reuse the learned information and reduce the amount of capacity that needs to be added. By using a ternary mask we can define three states:
\begin{itemize}
    \item \textbf{forward only}: features are used during the forward pass so that the learned information from previous tasks is used; but the backward pass step is removed in order to keep the weights and prevent forgetting. This state is used on features from previous tasks.
    \item \textbf{normal}: forward and backward passes are applied as usual to learn the task at hand. This state is used on new features created by the network expansion.
    \item \textbf{masked}: neither forward nor backward passes are allowed, the features do not contribute to the network inference and the weights associated to it are frozen. This state is used at test time only when evaluating an old task after a new task is added. When extending the capacity of the network, the new features will not be used when doing inference on the previous tasks since those did not exist at the moment of their training.
\end{itemize}
Similar as in the case of the binary mask we assign features to tasks with a mask $m^{t,l}$ (with $l$ the corresponding layer). Again overlap in the selected features is not allowed. However, different than before, we now define a second mask $n^{t,l}$ per task $t$ which is defined as:
\begin{equation}
    n^{t,l}_{i} = 
    \begin{cases}
        1,  & \text{if } \exists\; s\leq t\;: m^{s,l}_{i}=1 \\
        0,  & \text{otherwise}
    \end{cases}
\end{equation}
The forward and backward pass are now given by:
\begin{equation}
\label{eq:forward_pass}
y=\left(Wx\right)\odot n^{t,l},
\end{equation}
\begin{equation}
\label{eq:backward_pass}
\frac{\partial \mathcal{L}}{\partial W_{ij}} = (n^{t,l-1}_j n^{t,l}_i - n^{t-1,l-1}_j n^{t-1,l}_i) x_j \frac{\partial \mathcal{L}}{\partial y_{i}},
\end{equation}
respectively. During the forward pass, features selected by previously learned tasks can be used in the current task. During the backward pass, we make sure that all new weights can be updated while forcing the existing ones from previous tasks to remain the same. In Fig.~\ref{fig:ternary_mask} we show an example with two tasks, where adding features to the layer allows for more connections to be used than in the binary case. The part of the mask corresponding to $n^{t,l-1}_j n^{t,l}_i$ corresponds to all available connections at task $t$. In a similar way, the part of the mask corresponding to $n^{t-1,l-1}_j n^{t-1,l}_i$ corresponds to all available connections at task $t-1$. Subtracting both terms allows us to mask the connections that contain the already learned content and apply backpropagation only on the new connections.

\figbinarymask  

Note that this definition also allows to use the same forward and backward pass in case we would want to re-train one of the previous tasks. However, since we do not contemplate this option for our proposed setup, we can simplify equation~\ref{eq:backward_pass} to (non-revisiting) task-IL. In this case, the forward pass remains the same as in equation~\ref{eq:forward_pass}, and the backward pass can be rewritten as:
\begin{equation}
\label{eq:backward_pass_2}
\frac{\partial \mathcal{L}}{\partial W_{ij}} = (m^{t,l}_i \ \vee\  m^{t,l-1}_j) x_j\frac{\partial \mathcal{L}}{\partial y_{i}}
\end{equation}
where $\vee$ is the OR logical operator which makes the mask active when either operands are active.

Since $n^{t,l}_{i}$ can never be 0 if one of the current or previous $m^{1..t,l}_{i}$ is 1, both masks $m^{t,l}$ and $n^{t,l}$ can be combined in a single ternary mask. This is because weights associated to a feature that is not used in the forward pass are never updated. With this ternary mask, the states are associated as follows: when $m^{t,l}_{i}\!=\!1$ and $n^{t,l}_{i}\!=\!1$ the neuron is used and learnable (normal state), when $m^{t,l}_{i}\!=\!0$ and $n^{t,l}_{i}\!=\!1$ the neuron is used and contributes to the forward pass but the associated weights are not updated (forward only state), and finally when $m^{t,l}_{i}\!=\!0$ and $n^{t,l}_{i}\!=\!0$ the neuron is unused, not taking part in the inference or the update of the network (masked state).

Allowing to use previously learned parameters in the forward pass, but only updating network parameters assigned to the current task in the backward pass is also applied in Packnet~\cite{mallya2018packnet} and HAT~\cite{serra2018overcoming}. However, in contrast to us, Packnet has the masks on the weights and not on the features. HAT applies a soft activation mask, which permits forgetting of previous tasks. We further distinguish from these methods by the task-specific feature normalization (discussed in the next section) which is a crucial ingredient of our method, and which allows not only to exploit previously learned features, but also to adapt them to the current task. This is not possible for neither Packnet nor HAT.

\figternarymask  

\subsection{Task-specific feature normalization (FN)}
Since binary or ternary masks freeze filters learned on previous tasks, those filters have no room for flexibility to small changes in the features. This means that even when being very similar to the ones needed for a new task, they tend to learn a similar version of those filters with shifted or scaled operators. This phenomenon is similar to the one observed when learning several styles for style transfer networks. A way of reusing learned filters in a more efficient way but still keep the zero-forgetting property would be to use a similar approach as \emph{conditional instance normalization}~\cite{dumoulin2017learned}, which consists in transforming a set of features $x$ into a normalized version $\hat{x}$ depending on the task.

Let $x_{l,1..I}$ be the features of layer $l$, and $\gamma_{t}$, $\beta_{t}$ the learnable parameters for each feature given a fixed task $t$. We define the task-specific feature normalization of $x_{l,i}$ as:
\begin{equation}
    \hat{x}_{l,i}\  =\  \gamma_{t,l,i} \  x_{l,i}\  + \ \beta_{t,l,i}
\end{equation}
where we apply a conditional normalization on the task without applying an instance normalization on the mean and standard deviation across the spatial dimensions. These parameters allow to slightly adjust the learned filters to the new tasks without modifying existing parameters (thus no forgetting happens) and with little overhead to the network capacity since the $\gamma$ and $\beta$ parameters are for each feature and not for all weights.

\subsection{Growing Ternary Feature Masks}
\label{sec:growing}
One of the core characteristics of our proposed method is that it can easily grow and expand the capacity of the network as is required. Given a network with $L$ layers, any layer with the corresponding $y_{l,1..I}$ learned features can be expanded if those learned features are not enough to represent the new task. When expanding a layer by $N$ new features, the output of the layer grows to $y_{l,1..I+N}$. That affects only the newly added forward mask values:
\begin{equation}
    n^{t,l}_{j} = 
    \begin{cases}
        1,  & \text{for current task, } \text{if } 1 \leq j \leq I+N\\  
        0,  & \text{for previous tasks, }  \text{if } I < j \leq I+N
    \end{cases}
\end{equation}
so that all features can be seen while learning the new task but ignored by previous tasks. Then, the backward mask:
\begin{equation}
    m^{t,l}_{j} = 
    \begin{cases}
        0,  & \text{for current task, }  \text{if } 1 \leq j \leq I \\
        1,  & \text{for current task, } \text{if } I < j \leq I+N \\
        0,  & \text{for previous tasks, }  \text{if } I < j \leq I+N
    \end{cases}
\end{equation}
so that it only affects the new connections without modifying previous knowledge.

Training small tasks on large networks at the beginning of a continual learning setup, usually leads to overfitting or too much repetition of filters. Feature usage on the new tasks look very unbalanced in comparison to learning larger tasks~\cite{masana2017domain}.
We believe that learning tasks in their correct capacity and growing when more is needed is a much better approach to avoid overfitting. This observation is backed by the better results some other approaches have when pruning and retraining on smaller sub-networks than when directly pruning or learning in larger sub-networks~\cite{mallya2018packnet, rusu2016progressive}.

\section{Experimental results}
\label{sec:exps}
In this section we report on a range of experiments to quantify the effectiveness of our proposed approach and compare with other state-of-the-art methods and baselines. More details can be found in the Supplementary material.\footnote{Code: {\scriptsize \url{https://github.com/mmasana/TernaryFeatureMasks}}}

\subsection{Experimental Setup}
\label{sec:exp_setup}
\minisection{Datasets.} We evaluate approaches on a larger lower resolution dataset (tiny ImageNet ILSVRC2012~\cite{deng2009imagenet}), on a large-scale dataset (ImageNet~\cite{russakovsky2015imagenet}) and some fine-grained classification datasets: Oxford 102 Flowers~\cite{nilsback2008automated}, CUB-200-2011 Birds~\cite{wah2011caltech} and Stanford Actions~\cite{yao2011human}. Statistics over those datasets are summarized in Table~\ref{table:datasets}. For all experiments we take a fixed random set of 10\% of images for validation. The validation set is equally distributed among the number of classes and fixed for each experiment to ensure a fair comparison. Since the test set is not labelled for ImageNet ILSVRC2012, we use the validation set for test instead.

\minisection{Network architectures.} For tiny ImageNet we use \mbox{VGG-16}, which provides high performance results~\cite{simonyan2014very}. Since tiny ImageNet has a low resolution, the last max-pool layer and the last three convolutional layers from the feature extractor are removed. For ImageNet and the fine-grained datasets we use AlexNet~\cite{krizhevsky2012imagenet}. The models are trained from scratch using only samples from train. Our proposed method TFM starts with a network that is smaller than the proposed ones at each layer (reduced number of output filters). Then, it grows as explained in Sec.~\ref{sec:growing} as more features are added every time a new task is learned. We limit the growth of the network to the total size of the one used by all other approaches.

\tabledatasets  

\minisection{Training details.} We train using backpropagation with plain Stochastic Gradient Descent following the setup of HAT~\cite{serra2018overcoming}. With a batch size of $64$, learning rate starts at $0.05$, decaying by a factor of $3$ when $5$ consecutive epochs have no improvement on the validation loss, until either the learning rate is reduced below $10^{-4}$ or 200 epochs have passed. Data splits, task sequence, data loader shuffle and network initialization are fixed for all approaches given a seed. Following~\cite{delange2021continual}, we use dropout with $p=0.5$.

\minisection{Baselines.} Finetuning uses the cross-entropy loss to learn each task as it comes, without using data from previous tasks nor avoiding catastrophic forgetting. Incremental Joint training breaks the no-revisiting data rule and learns with data from the current task as well as all the previous tasks, serving as an upper-bound to compare all approaches. Finally, we propose to use Freezing as a baseline where we learn the first task and then freeze all layers except the head for the remaining tasks.

\minisection{Hyperparameters.} Distillation and model-based approaches use hyperparameters to control the trade-off between forgetting and intransigence on the knowledge of previous tasks. On top of that, LwF has a temperature scaling hyperparameter for the cross-entropy loss. From the mask-based models, HAT has a trade-off hyperparameter too and a maximum for the sigmoid gate steepness. PackNet has a prune percentage of the layers.

For TFM, at each new task, several growth percentages are evaluated on the validation set without the knowledge of previous or future tasks. We pick the lowest growth rate which obtains a performance within a margin of the best performance (we set the margin to be 1.5\% for tiny ImageNet and 0.1\% for fine-grained). Then, we learn the task at hand on train and move to the next task. For ImageNet this scheme would be computationally demanding and we use a fixed growth schedule, starting from 55\% of the weights for the first task and add 5\% for all remaining tasks.

\subsection{Fine-grained datasets}
A common setup to evaluate task-IL over a number of learning sessions are disjoint splits (tasks) inside the same classification dataset. It should be noted that we start training from scratch resulting in lower scores than reported by papers which train from a pretrained network. However, because of the large number of classes in ImageNet (including a subset of Birds) we consider training from scratch provides a more natural setting for continual learning.

We compare our method (TFM) and an ablation version of it without the task-specific feature normalization (TFM w/o FN) with two mask-based approaches (HAT, PackNet), a well-known model-based approach (EWC) and the baselines (Finetuning, Freezing, Joint) on three fine-grained datasets (Flowers, Birds, Actions). As can be seen in Table~\ref{tab:fine_grained}, our approach outperforms the other approaches for the three datasets. For these datasets only on Flowers a considerable performance gain is observed when adding task-specific feature normalization. Only PackNet manages to obtain competitive results, however, on both Birds and Actions, TFM does significantly better, while having a much lower memory overhead than PackNet (0.2Mb versus 27.3Mb respectively). It is also interesting to note how well Freezing works as a non-forgetting baseline.

\finegrainedtable{Comparison with the state-of-the-art. Accuracy after learning 4 tasks on AlexNet from scratch. Number between brackets indicates forgetting.}

\subsection{Task-similarity effects on tiny ImageNet}
Next we experiment on several ten-task splits of tiny ImageNet. We compare our approach (TFM) with two distillation methods (LFL~\cite{jung2016less}, LwF~\cite{li2017learning}), two model-based methods (EWC~\cite{kirkpatrick2017overcoming}, IMM~\cite{lee2017overcoming}) and two mask-based methods (HAT~\cite{serra2018overcoming}, PackNet~\cite{mallya2018packnet}). We also include two baselines (Finetuning, Freezing). Performance is evaluated under the same conditions on a random tiny ImageNet partition and on a semantically similar partition (see Table~\ref{tab:tiny_imagenet_all}). For further information on the latter and for per-task comparison, check the Supplementary material.

For random splits most methods have quite good results on the last tasks with minor to no forgetting. LFL, IMM and EWC provide some improvement over Finetuning. LwF has a very good performance due to tasks being quite similar. All mask-based models have a very similar performance, with PackNet having the better performance. In the semantically similar splits, which has a more different distribution for each task than the random case, some approaches have difficulties avoiding catastrophic forgetting as the sequence gets longer. It is interesting to see, that the good performance of LwF on the random split is not transferred when we using semantic splits. As observed before~\cite{aljundi2016expert}, LwF fails when there exists large changes in the distributions between tasks. Mask-based models outperform all other approaches again, with TFM having the better performance. Freezing the feature extractor after the first task and learning only the classifier for the remaining tasks works better in the semantically similar splits than in the random splits.

\tinytable{Comparison with Tiny ImageNet on VGGnet from scratch. Average accuracy after learning all tasks. Classes are randomly split and fixed for all approaches.}
\imagenettable{Comparison with the state-of-the-art. ImageNet on AlexNet from scratch. Accuracy of each task after learning all tasks. Number between brackets indicates forgetting.}

\subsection{Effect of starting-task size on tiny ImageNet}
We also propose an experimental setup where the first task of tiny ImageNet uses 110 classes (55\%) while the remaining 9 tasks use 10 classes (5\%) each (see Table~\ref{tab:tiny_imagenet_all}). This allows most of the methods to start with a rich representation after learning the first task. In this setup, comparing existing methods with the Freezing baseline is more interesting. EWC shows little forgetting, but that causes the model to become too rigid and learn the rest of the tasks with more difficulty and having a lower overall performance. Trying to lower the trade-off hyperparameter shows a stronger forgetting of the first tasks and causes severe catastrophic forgetting. Distillation approaches try to keep representations the same as new tasks are learned. However, small changes in the weights cause forgetting later into the sequence. HAT works fine, but with a limited capacity to make changes, ends up not learning the new tasks as easily. Freezing the network after the first task seems to be one of the best options in this setup, since the rich representation of the first 110 classes is a good starting point to learn the rest of the tasks with a simple classifier. We therefore advocate for the Freezing baseline to be included in continual learning comparisons since it often provides a much harder baseline than Finetuning. Only PackNet and TFM are able to improve over that baseline even if they start from a smaller capacity, with TFM having the best results. We again refer to the Supplementary material for further per-task results.

\subsection{ImageNet}
\label{sec:imagenet_exp}
Most of the compared task-aware approaches have not been evaluated using a large-scale dataset such as ImageNet. We therefore compare our proposed method (TFM) with some of those state-of-the-art approaches. In Table~\ref{tab:exp_imagenet} we can see that TFM outperforms all other approaches on ImageNet split into 10 tasks of random classes. LwF does well when learning each new task with the help of the representations of previous tasks. However, as more tasks are included, older tasks start forgetting more. IMM (mode) has the opposite effect, it focuses on intransigence and tries to keep the knowledge of older tasks, running out of capacity for the newer tasks. This allows for the approach to not forget much and even have a small backward transfer, but at the cost of performing worse with newer tasks. EWC has the worst performance, possibly due to the difficulty of having a good approximation of the FIM when there is so many classes per task. HAT had problems scaling to this scenario, showing difficulties to learn new tasks. Both PackNet and TFM have a good overall zero-forgetting performance, and rely on the amount of capacity of the network more than other approaches. PackNet has a better performance during the first three tasks, taking advantage of the compression power of the pruning and finetuning. TFM has much less capacity for those tasks and therefore provides a bit lower start. However, as the remaining capacity of the network gets smaller for PackNet, TFM is capable of growing at a more scalable pace, getting a better performance on the remaining seven tasks and achieving the best results overall.

\section{Conclusions}
\label{sec:conclusions}
For many practical applications, it is important that network accuracy on tasks does not deteriorate when learning new tasks. Therefore, in this paper, we propose a new method for continual learning which does not suffer from any forgetting. Other than previous methods which apply masks to the weights, we propose to move the mask to the features (activations). This greatly reduces the number of extra parameters which are added per task and reduce the overhead of the network in which other approaches incur. In addition, we propose to apply a task-specific feature normalization of features, which allows adjusting previously learned features to new tasks. In ablation experiments this was found to improve results of the ternary feature masks. Furthermore, when compared to a wide range of other continual learning techniques, our method consistently outperforms these methods on a variety of datasets. However, as a limitation, the usage of mask-based approaches becomes computationally less efficient when moving from the proposed task-IL scenario to a class-incremental one. The absence of the task-label at inference time requires mask-based approaches to evaluate one forward pass per task to provide the joint prediction. We consider adapting mask-based approaches to a class-incremental setting as an interesting direction for future research.

\section*{Acknowledgments}
We would like to thank Biel Villalonga and Mikel Menta for their helpful discussion. Marc Masana acknowledges the \mbox{2019-FI\_B2-00189} grant from Generalitat de Catalunya. We acknowledge the financial support by the Spanish project PID2019-104174GB-I00. We also thank KU Leuven C1 project Macchina.

{\small
\bibliographystyle{ieee_fullname}
\bibliography{egbib}
}

\newpage
\renewcommand{\thetable}{S\arabic{table}}
\renewcommand{\thefigure}{S\arabic{figure}}
\renewcommand{\thealgorithm}{S\arabic{algorithm}}

\appendix
\section*{Appendix}

\section{More on related work}
We show in Table~\ref{tab:soa_comparison} a comprehensive overview of some of the characteristics that we consider to be more relevant to the experimental setup we propose.
\soacomparisontable

\section{Ternary mask implementation}
The formulation of our proposed method in Sec.~3 states that we can combine both masks $m^t$ and $n^t$ into a ternary mask. In order to make the implementation easier and more efficient, this is done by creating a ternary mask that is set to be state $2$ for all features at task 1. This means that the first task will work as a normal network that allows for learning the task at hand as if we were using finetuning. Then, when moving to task 2, we will grow the network and add new features. The masks for task 1 associated with the new features will be set to state $0$, therefore these are not used when evaluating nor are they learnable for task 1. The masks for task 2 will then be created by setting the previous existing features to state $1$ and the new added features to state $2$. This would allow the features with state $1$ to be used during the forward pass for mask $n^{t=2}$, while the features with state $2$ will contribute to both forward pass for mask $n^{t=2}$ and backward pass for mask $m^{t=2}$ (using Eq.~6). This process is explained in Algorithm~\ref{table:algorithm}.

\section{Experimental setup details}
\minisection{Datasets.} Tiny ImageNet is a $64\times64\times3$ resized version of 200 ImageNet classes. ImageNet uses $256\times256\times3$ inputs that use random cropping to $224\times224\times3$ during training for data augmentation. In the case of Birds we resize the bounding box annotations of the objects to $224\times224\times3$ for all splits. We do the same for Actions but without using the bounding box annotations. For Flowers we resize to $256\times256\times3$ and also do data augmentation by random cropping $224\times224\times3$ patches during training and using the central crop for evaluation. In all experiments we perform random horizontal flips during training for data augmentation.

We decide to not do experiments on permuted MNIST since it has been shown to not allow fair comparison between different approaches~\cite{lee2017overcoming}. The MNIST data contains a too large amount of zeros per input, which leads to an easy identification of important weights that can be frozen to not overlap with the other tasks. Furthermore, the MNIST data might be too simple to represent more realistic scenarios.

\minisection{Hyperparameters.} Distillation and model-based approaches use hyperparameters to control the trade-off between forgetting and intransigence on the knowledge of previous tasks. On top of that, LwF has a temperature scaling hyperparameter for the cross-entropy loss. From the mask-based models, HAT has a trade-off hyperparameter too and a maximum for the sigmoid gate steepness. PackNet has a prune percentage of the layers. We consider the values proposed by the papers if they have results on the same experimental setup. Otherwise, we use the hyperparameter search proposed in~\cite{delange2021continual} and~\cite{masana2020class}, which chooses hyperparameters with only the information at hand for each task.

\growingalgorithm

\section{A note on choosing expansion rates}
\label{sec:appendixA}
The continual learning philosophy states the rule of not using data from previous tasks when learning new ones; only data of the current task can be used at each step of the setup. It is also common in machine learning setups to use a part of the training set as validation in order to choose the best hyper-parameters. Therefore, when learning each task, a validation set of that specific task at hand can be used to train the network avoiding overfitting, but no other data can be used (neither from test nor from other previous or future tasks). It is important to state that we strictly comply to these rules in the experiments we propose.

As explained in Section~3.4, our proposed approach can be expanded as needed in order to learn the new tasks without having to change the connections from previous tasks. However, this flexibility of choosing how many features will be added to each layer can easily become a rabbit-hole of architecture optimization. Because of that, we decide to propose a simple setup for how we apply our proposed approach to be comparable to the other state-of-the-art. We take the maximum layer size for all approaches to be the same as the VGGnet or AlexNet architectures for the experiments in Section~4. This way, all approaches will have a similar number of parameters.

The differences between Figures~2 and~3 are further explained in Fig.~\ref{fig:3tasks}. During the first task, the only used features are those that have a grey-border in them. This means that for the two shown layers, task 1 is learned by using 8 features (with 12 connections). Once task 2 arrives, we fix the grey-border features and expand the network with the green-border ones. The new task then uses the existing network and expands it with 5 features (with 24 new connections). The masks for the green-border features are set to unused for masks of task 1, while they are learnable for task 2. Finally, as task 3 comes, 5 features are also added (with 36 new connections). Masks corresponding to the new features are set to unused for tasks 1 and 2, while set to learnable for task 3. This way of expanding the network also shows that as we learn more tasks, more knowledge is available from previous ones. This opens the possibility of having to add less and less features over time since the addition of each feature creates more connections to learn. In practice, one can imagine that when learning new tasks that are very similar to previous ones, no new features will have to be added and the current network knowledge and a specific head for the task will be enough. Further research and analysis on the specific details for each layer expansion and architecture is left for future work.

\section{Tiny ImageNet semantic splits}
\label{app:semantic}
The semantically similar splits of tiny ImageNet used in Table~4 experiments are grouped as described in Table~\ref{tab:tiny_group_split_classes}.

\section{Extended results for Tiny ImageNet}
We present in Tables~\ref{tab:tiny_imagenet_random},~\ref{tab:tiny_imagenet_groups} and~\ref{tab:tiny_large_task} more detailed accuracy and forgetting results for the experiment on Tiny ImageNet with random split, semantic split and larger first task, respectively. Both Tables~\ref{tab:tiny_imagenet_random} and~\ref{tab:tiny_imagenet_groups} have a sequence of 10 tasks with 10 classes each, but with different class orderings. Table~\ref{tab:tiny_large_task} shows results when starting with a larger first task, and thus the last column is only averaged over the smaller tasks (T2-T10). Results show that mask-based approaches achieve a better overall performance than other approaches on all splits, with TFM having the best performance or tied with the best.
\ternarythreetasks

\tinygroupsplits{0.45\textwidth}
\multipledatasets{20 0 0 20}{Comparison on a sequence of multiple datasets on AlexNet from scratch.}

\section{Multi-dataset experiment}
To further explore the performance of our proposed method when the tasks have a much different distribution, we shows results on a sequence of multiple fine-grained datasets: Oxford Flowers~\cite{nilsback2008automated}, MIT Indoor Scenes~\cite{quattoni2009recognizing}, CUB-200-2011 Birds~\cite{wah2011caltech}, Stanford Cars~\cite{krause20133d}, FGVC Aircraft~\cite{maji2013fine}, and Stanford Actions~\cite{yao2011human}. We compare TFM with the Finetuning baseline, EWC and HAT on AlexNet trained from scratch. Each of the approaches seems to have a tendency on doing better in different tasks. However, TFM is never the worse at any task and has a clear better average accuracy after learning all tasks. This seems to indicate that our proposed method can still perform quite well even when there are larger distribution changes between tasks.

\tinyrandomtable{Comparison with the state-of-the-art. Tiny ImageNet on VGGnet from scratch. Accuracy of each task after learning all tasks. Number between brackets indicates forgetting. Classes are randomly split and fixed for all approaches.}
\tinygroupstable{Comparison with the state-of-the-art. Tiny ImageNet on VGGnet from scratch. Accuracy of each task after learning all tasks. Number between brackets indicates forgetting. Classes are split by semantic closeness and fixed for all approaches.}
\tinylargetask{Comparison with the state-of-the-art. Tiny ImageNet on VGGnet from scratch. Accuracy of each task after learning all tasks. Numbers between brackets indicates forgetting. LArger first task. Average on the smaller tasks 2 to 10.}

\end{document}